\documentclass[preprint,12pt,authoryear]{elsarticle}

\usepackage{caption}
\usepackage{amssymb}
\usepackage{diagbox}
\usepackage{multicol}
\usepackage{multirow}
\usepackage{booktabs}
\usepackage{array}
\usepackage{makecell}
\usepackage{multirow}
\usepackage{pdflscape}
\usepackage{rotating}

\journal{Computer Vision and Image Understanding}

\begin{document}

\begin{sloppy}  

\begin{frontmatter}



\title{FAR-AMTN: Attention Multi-Task Network for Face Attribute Recognition}





\author[1]{Gong Gao}
\author[1]{Zekai Wang}
\author[1]{Xianhui Liu}
\author[1]{Weidong Zhao}
\affiliation[1]{organization={School of Electronic and Information Engineering, Tongji University},
                city={Shanghai},
                postcode={200092},
                state={Shanghai},
                country={China}}
\affiliation[2]{organization={School of Electronics and Computer Science, University of Southampton},
            city={Southampton},
            postcode={SO17 1BJ}, 
            state={Southampton},
            country={United Kingdom}}

\cortext[1]{Xianhui Liu}

\begin{abstract}
To enhance the generalization performance of Multi-Task Networks (MTN) in Face Attribute Recognition (FAR), it is crucial to share relevant information across multiple related prediction tasks effectively. Traditional MTN methods create shared low-level modules and distinct high-level modules, causing an exponential increase in model parameters with the addition of tasks. This approach also limits feature interaction at the high level, hindering the exploration of semantic relations among attributes, thereby affecting generalization negatively. In response, this study introduces FAR-AMTN, a novel Attention Multi-Task Network for FAR. It incorporates a Weight-Shared Group-Specific Attention (WSGSA) module with shared parameters to minimize complexity while improving group feature representation. Furthermore, a Cross-Group Feature Fusion (CGFF) module is utilized to foster interactions between attribute groups, enhancing feature learning. A Dynamic Weighting Strategy (DWS) is also introduced for synchronized task convergence. Experiments on the CelebA and LFWA datasets demonstrate that the proposed FAR-AMTN demonstrates superior accuracy with significantly fewer parameters compared to existing models.
\end{abstract}


\end{frontmatter}

\section{Introduction}
Face Attribute Recognition (FAR) is frequently employed as an auxiliary task in various applications, including face recognition \citep{iranmanesh2018deep,pattnaik2023face}, face retrieval \citep{zaeemzadeh2021face,maroto2023active}, and face editing \citep{kou2023character,yang2023dfsgan,liu2023sketch2photo}, among others. Given the extensive array of facial attributes, Multi-Task Networks (MTN) \citep{li2015multi} are often utilized for feature extraction. MTN, a paradigm in machine learning, seeks to enhance the performance of interrelated tasks by capitalizing on valuable information from each. By grouping related tasks \citep{chen2021improving,song2022prior,chen2023learning}, MTN effectively streamlines FAR, thereby reducing the parameter count. 
However, the absence of feature interaction among non-shared, high-level modules may impair generalization performance.
Moreover, the variable convergence rates and loss scales across tasks in MTN necessitate multiple adjustments of hyper-parameters to achieve the global optimal solution, consequently extending the training duration.

FAR research encompasses various methodologies, broadly categorized into multi-label learning approaches and multi-task learning approaches.
Multi-label learning methods directly produce prediction vectors, which are then refined using SVM or thresholds for the final predictions.
In 2015, \citet{liu2015deep} introduced a two-module approach for FAR: Localization Networks (LNets) for face localization and Attribute Network (ANet) for feature extraction, with SVM for attribute recognition subsequently.
\citet{sharma2020slim} developed Slim-CNN, employing depthwise separable convolutions merged with pointwise convolutions, culminating in 40 vectors through a fully connected layer. 
\citet{zheng2022general} proposed the FaRL framework, which leverages large-scale data for pre-training, performs weighted fusion on three feature layers to obtain global features, and ultimately generates 40 vectors through a fully connected layer.
Although these methods enhance prediction efficiency, they do so at the expense of performance.

Multi-task learning methods utilize grouping strategies \citep{chen2021improving,song2022prior,chen2023learning} to learn heterogeneous tasks and weighted loss functions \citep{huang2008labeled,rudd2016moon,wang2019dynamic,du2021parameter,lingenfelter2021improving,serna2022sensitive} to improve the sensitivity towards certain tasks.

When integrating Face Attribute Recognition (FAR) with multi-task learning, three primary challenges emerge.
Firstly, as illustrated in Fig. \ref{fig:1}, employing a multi-task framework leads to a significant increase in the parameters with the addition of tasks. Secondly, tower-like Multi-Task Networks
\citep{mao2020deep,chen2023learning}, which lack group feature fusion, exhibit high similarity in the distribution of parameters across independent modules, thereby limiting their generalization capabilities. Lastly, Fig. \ref{fig:1} highlights the disparity in loss scale and task-specific gradient descent rates during MTN training.
This discrepancy can result in a seesaw effect \citep{zhang2021deep} and task-specific overfitting \citep{park2021influence,santos2018cross}, complicating the optimization process.

\begin{figure}[t]
\centering
\includegraphics[width=1.0\linewidth]{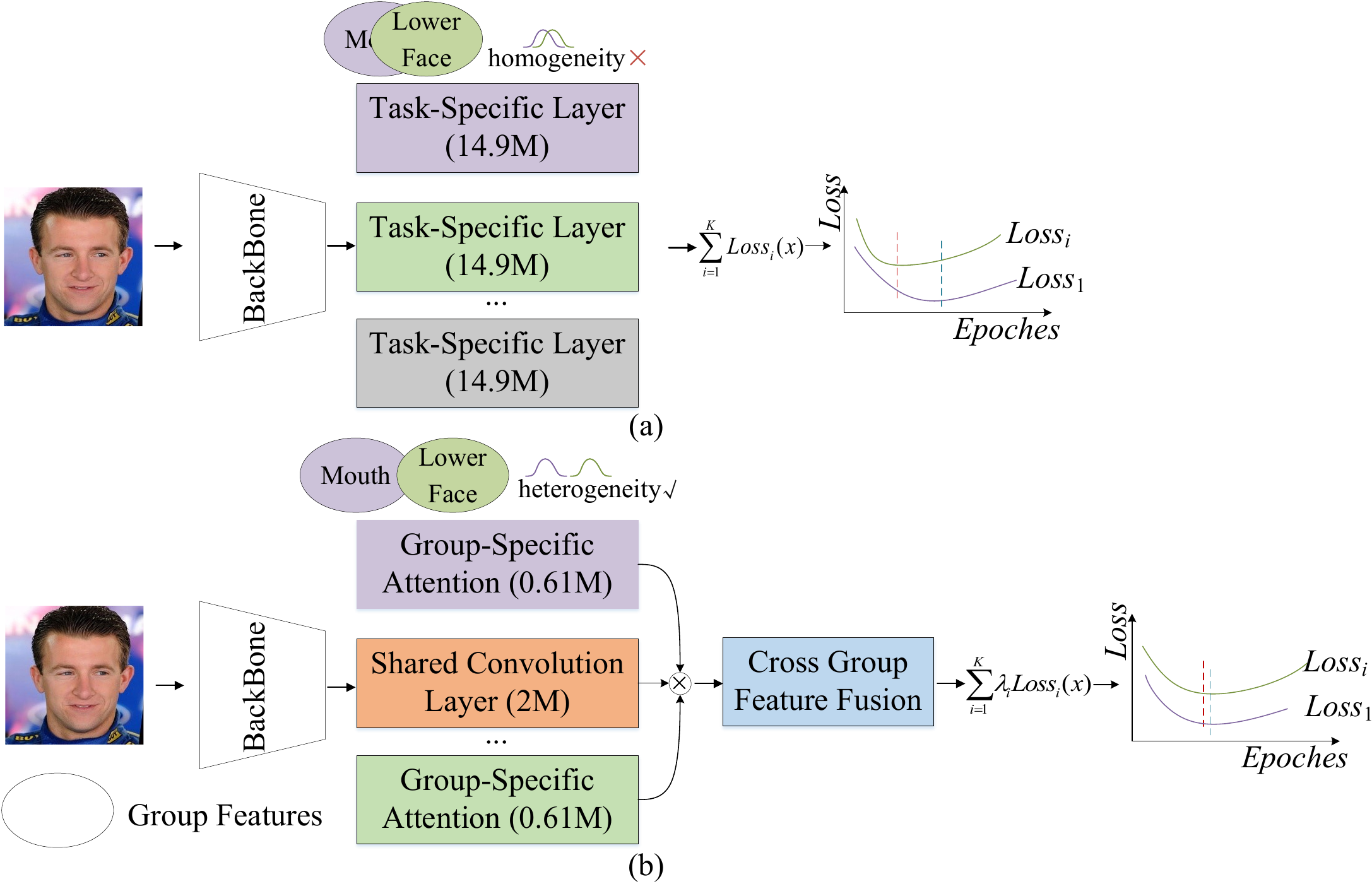}
\caption{(a) The preceding methodology of multi-task  attribute recognition. (b) The FAR-AMTN method. To address three issues in existing methods: the increase in model parameters with the number of tasks, the heterogeneity of group features, and asynchronous convergence, we propose specific group attention layers with parameter sharing mechanism and cross-group feature fusion modules. Additionally, we enhance the dynamic weighting $\lambda$ strategy to enable synchronous convergence for multiple tasks. Where $\otimes $ denotes the element-wise multiplication of two elements.}
\label{fig:1}
\end{figure}

To address the challenges outlined above, we introduce FAR-AMTN, incorporating both the Weight-Shared Group-Specific Attention (WSGSA) module and the Cross-Group Feature Fusion (CGFF) module, as shown in Fig. \ref{fig:1}(b). The WSGSA module utilizes a shared convolutional parameter to enhance group feature representation, thereby not only improving the characterization of group features but also reducing the parameter count of the Multi-Task Network (MTN) model. The CGFF module, on the other hand, explores and leverages the relationships between group attributes by facilitating interaction among group features. Additionally, we employ a Dynamic Weighting Strategy (DWS) for network training, which mitigates overfitting issues arising from variable gradient descent rates and loss scales among tasks. FAR-AMTN demonstrates superior performance on the CelebA \citep{liu2018large} and LFWA \citep{huang2008labeled} datasets. The contributions of this work are summarized as follows:

\begin{itemize}
\item Unlike traditional tower-like Multi-Task Networks, FAR-AMTN incorporates the Weight-Shared Group-Specific Attention (WSGSA) module, utilizing shared parameters to improve group feature representation and simultaneously reduce the overall parameter count of the MTN. 

\item FAR-AMTN facilitates feature interaction through the Cross-Group Feature Fusion (CGFF) module, enabling the exploration and utilization of the mutual relationships among group features. The Dynamic Weighting Strategy (DWS) is implemented to achieve synchronized task convergence by addressing both the gradient descent rate and the loss scale.

\item Testing on the CelebA and LFWA datasets has shown that FAR-AMTN not only achieves superior performance but also maintains a lower parameter count in comparison to existing multi-task learning approaches.
\end{itemize}

\section{Related Works}

\subsection{Multi-Task Network and Face Attribute Recognition}

{\bf Multi-Task Network.}
Numerous computer vision applications require real-time solutions for multiple tasks. Employing multi-task learning to train a neural network facilitates the simultaneous handling of multiple tasks while minimizing the parameter count. The set of tasks can be represented as $T = \{Task_i\}, i = 1, 2, ..., K$, and the comprehensive optimization of the MTN can be formulated as
\begin{equation}
    \min \sum\limits_{i=1}^K {L({\mathop{\rm I}\nolimits} ,{y_i},{w_i})},
\end{equation}
where $K$ denotes the number of tasks, ${\rm{I}} \in {{\rm{R}}^{{H}\times W}}$  represents the network input, with $H$ and $W$ indicating the dimensions of the input matrix. The variable ${y}$ signifies the label vector, and ${{\emph{w}}_i} $ denotes the weight parameter for the \emph{i}-th task.
Given the assumption that the tasks are interrelated, the weight parameters for each task can be concatenated to derive the collective parameters.

However, due to the inherent relationships among tasks, directly utilizing features from MTN does not yield optimal performance. Consequently, many researchers have explored task interaction representations to investigate task relationships, such as employing gated multi-expert models \citep{tao2023smoke,ye2023taskexpert} or task interaction models \citep{ye2022taskprompter,vandenhende2020mti,ye2022inverted,xu2022mtformer}. \citet{ye2023taskexpert} and \citet{tao2023smoke} proposed a mixture-of-experts model that utilizes expert networks and task-specific modules to provide more discriminative representations for different tasks. The above approaches employed a gating mechanism to control the importance of various features. Nevertheless, the absence of inter-task interaction modeling may lead to the learning of shared features in scenarios with multiple attribute groups, potentially affecting performance.
\citet{ye2022taskprompter} proposed a novel spatial-channel multi-task prompting transformer framework called TaskPrompter, which models task-generic and task-specific representations as well as cross-task interactions in an end-to-end manner, leveraging attention mechanisms to enhance representation learning for dense prediction tasks. 
\citet{vandenhende2020mti}, \citet{ye2022inverted}, and \citet{xu2022mtformer} proposed a task interaction module based on MTN, which effectively transmits refined task information using attention mechanisms while aggregating fine-tuned task features, ultimately generating predictions for each task. The task interaction model has shown significant improvements in performance across tasks such as depth estimation, semantic segmentation, and instance segmentation. Nevertheless, due to the large number of tasks and their complex relationships in FAR, there is a need for specifically designed effective feature interaction modules. Additionally, the parameter count of these methods tends to increase significantly with the number of tasks, which complicates model training.

{\bf Face Attribute Recognition based on Multi-Task Learning.}
Deep learning has significantly contributed to the advancement of feature learning for FAR. This section provides an overview of methods that employ multi-task learning for FAR.

Researchers have incorporated auxiliary tasks such as identity information  \citep{cao2019learning,yang2020hierarchical}, facial organ location information \citep{ehrlich2016facial,ding2018deep,10064142}, and semantic location information \citep{kalayeh2017improving,zheng2022general} to refine feature representation and improve FAR performance.
 \citet{cao2019learning} proposed a Partially Shared Multi-task CNN (PS-MCNN) to learn both shared and task-specific features, using identity information to enhance the performance of face attribute estimation.
 \citet{yang2020hierarchical} developed a Hierarchical Feature Embedding (HFE) framework that integrates attribute and ID information for fine-grained feature embedding.
  
 \citet{ding2018deep} proposed a cascade network that learns to localize attribute-specific facial regions and perform attribute classification without alignment.
  \citet{kalayeh2017improving} employed a deep semantic segmentation network to create a face attribute prediction model. \citet{10064142} proposed an Identity-aware Contrastive Knowledge Distillation (ICKD), which uses three different knowledge distillation schemes at different distillation points, namely, feature-based, relation-based and response-based distillation loss. However, the use of auxiliary networks for joint training has shown limited relevance and occasionally resulted in negative transfer. Additionally, these auxiliary tasks require extra supervisory signals, rendering them impractical for FAR in uncontrolled environments.

Consequently, many researchers have focused on directly applying MTN for FAR. \citet{lu2017fully}, \citet{mao2020deep}, \citet{chen2021improving}, and \citet{tao2024hierarchical} introduced a multi-task architecture that extracts features using a backbone network with a feature fusion layer and a task-specific attention layer, employing a grouping strategy to leverage prior knowledge of attribute relationships. 
Moreover, \citet{chen2023learning} proposed an Attention-aware Parallel Sharing (APS) network that adaptively extracts crucial features from each block of the shared sub-network by capitalizing on shared low-level features and task-specific sub-networks. \citet{tao2024hierarchical} proposed a hierarchical attention network with a progressive feature fusion strategy that effectively enhances the representational features of key facial regions while suppressing distracting features by integrating diverse feature extraction and hierarchical attention modules. While these methods utilize non-shared high-level modules to learn heterogeneous features, this strategy significantly increases the number of model parameters and the absence of high-level feature interaction compromises generalization capability.

\subsection{Dynamic Weighting Loss}
During the training phase of MTN, variable gradient descent rates and loss scales across tasks can prevent synchronous task convergence \citep{lv2022synchronous}, leading to the overfitting of certain tasks.
To address this, hyper-parameters are often employed to adjust the loss weight of each task, facilitating synchronous convergence.
\citet{mao2020deep}, \citet{hand2018doing}, and \citet{wang2019dynamic} have suggested updating the weights of each task based on the loss magnitude.
Consequently, the aggregate loss function of the MTN can be formulated as
\begin{equation}
    L=\sum\limits_{\emph{i}=1}^K {{\lambda _i}L({\mathop{\rm I}\nolimits} ,{y_i},{w_i})}, 
\end{equation}
where $\lambda$ denotes the relative importance weighting parameter for tasks. Throughout the training process, the magnitude of the gradient and the rate of convergence for each task dynamically fluctuate, making the adjustment of task weighting parameters a complex endeavor.

 To achieve synchronous convergence in multi-task learning, numerous researchers have proposed using the speed of gradient descent as a weighting parameter for tasks.
 \citet{he2017adaptively} and \citet{serna2022sensitive} introduced a dynamic weighting loss function, which is tailored to dynamically update the weight of each task during the training process. The adjustment of task weights is based on the magnitude of the validation set loss, leading to the derivation of a weighted loss as the aggregate loss in the training phase. \citet{liu2019end} developed a Dynamic Weight Average (DWA) algorithm, positing that tasks with quicker gradient descent rates should receive higher weight allocations. Although these methods assign weighted values based on the gradient descent speed of various tasks, they tend to neglect the scale of the loss.

\section{Method}
\subsection{Network Architecture}
Developed by Microsoft Research, ResNet \citep{he2016deep} represents a groundbreaking convolutional neural network that has seen extensive application in FAR. Within the FAR-AMTN model, ResNet50 serves as the foundational backbone network.
The  conv1 through conv4 layers of ResNet50, are utilized as shared lower layers to extract common low-level features. Subsequently, the final three convolutional layers of conv5 (1×1, 3×3, and 1×1) are augmented with the WSGSA module. 
Leveraging the WSGSA module to obtain enhanced group features, the CGFF module is then employed for facilitating feature interaction, enabling the modeling of complex relationships among different group attributes. 
Ultimately, attribute prediction outcomes are generated using $K$ distinct fully connected layers. Overall, the FAR-AMTN architecture synergizes the capabilities of ResNet, the WSGSA module, and the CGFF module to deliver precise and robust FAR performance.
\begin{figure*}[!htbp]
   \begin{center}
    \includegraphics[width=1.0\linewidth]{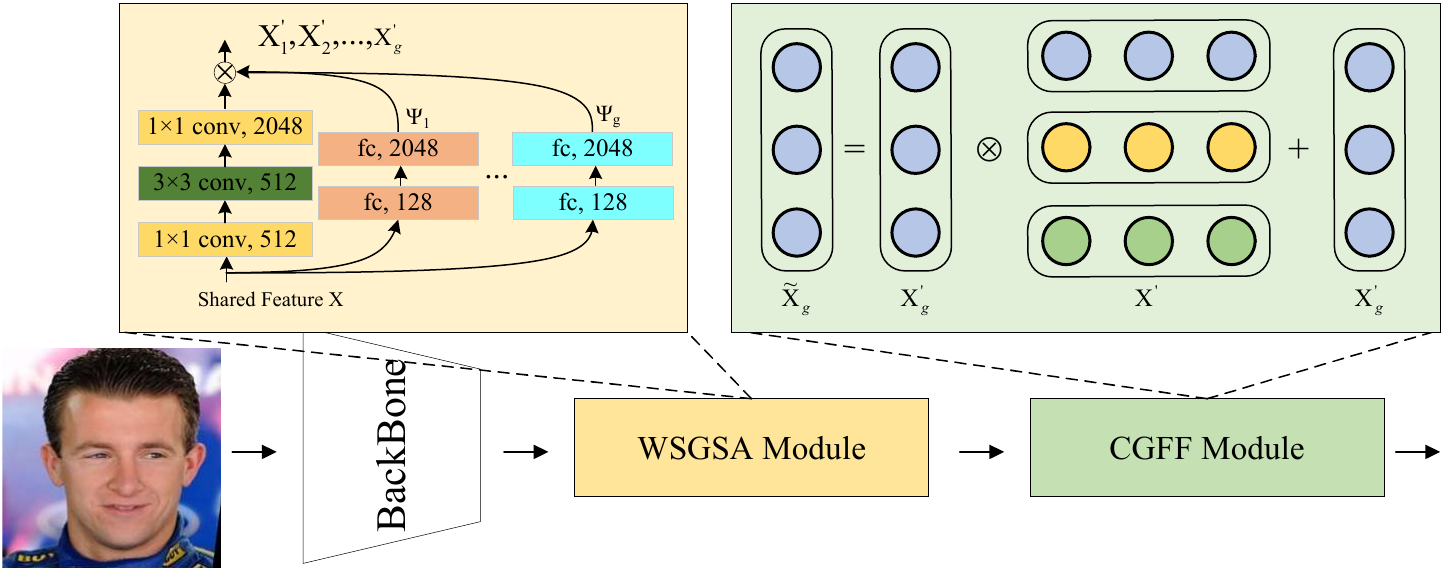}
      \caption{FAR-AMTN architecture diagram. The proposed FAR-AMTN architecture consists of Shared Bottom module, Weight-Shared Group-Specific Attention (WSGSA) module, and Cross-Group Feature Fusion (CGFF) module. In the diagram, $\otimes$ represents the element-wise multiplication of two elements. The dynamic weighting strategy is in the section \ref{sec:Dynamic Weighting Strategy}.}
      \label{fig:24}
   \end{center}

   \end{figure*}

\subsection{Weight-Shared Group-Specific Attention }

To enhance the representation of group features without a substantial increase in the number of parameters, we substitute the final Resblock of conv5 with the WSGSA module. This module employs a shared-parameter WSGSA to extract attention for each group. Following this modification, to compact the feature representation spatially, mean pooling processing is employed, which can be formulated as
\begin{equation}
    \rm{Z} = \frac{1}{{\emph{H} \times \emph{W}}}\sum\limits_{\emph{i} = 1}^\emph{H}{\sum\limits_{\emph{j} = 1}^\emph{W} {\rm{X}}}(\emph{i},\emph{j}),
\end{equation}
where $\rm{X}$ denotes the input of the WSGSA module. Upon obtaining the compression feature $\rm{Z}$, each group constructs two fully connected layers to learn heterogeneous group attention. The aggregated channel attention can be denoted as ${\rm{\Psi  = \{ }}{{\rm{\Psi}}_g}{\rm{\},}}g{\rm{ = 1,2,}}...G$, where ${\rm{\Psi}}$ is unique to each group in the GSA module. Here, $\emph{G}$ indicates the number of groups, and the group-specific channel attention ${\rm{\Psi}}_g \in {{\rm{R}}^{1 \times C}}$ can be formulated as
\begin{equation}
    {{\rm{\Psi }}_g} =Sigmoid(fc_{g,C\rightarrow C/16}(ReLU{\rm{(}}fc_{g,C/16\rightarrow C}{\rm{(Z))}})),
\end{equation}
where $fc_{g,C\rightarrow C/16}$ denotes a fully connected layer with number of nodes is reduced from $C$ to $C/16$. Conversely, $fc_{g,C/16\rightarrow C}$ signifies an increase in the number of nodes from $C/16$ back to $C$, aligning with the channel dimensions of the shared feature $\rm{X}$. $\emph{ReLU}$ and $\emph{Sigmoid}$ indicate the activation function. The feature of each group can be formulated as 
\begin{equation}
{\rm{X}}_{\emph{g}}^{'}= {\rm{\Psi} _{\emph{g}}} \otimes \emph{f}({\rm{X}}),
\end{equation}
where $\emph{f}$ represents the convolution operation, which is shared in the WSGSA module.

\subsection{Cross-Group Feature Fusion}
In the field of research, it is common for researchers to directly utilize group features for predictions or to employ methods such as concatenation and addition. However, this approach does not fully account for the heterogeneity inherent in group features. Given that each group is representative of a unique facial area, the features of these groups are inherently heterogeneous. Therefore, it is imperative to engage in feature interaction to elucidate the nonlinear relationships among these features.

After obtaining each group feature ${{\rm{X}}_g^{'}}$, we use CGFF to carry out feature interaction, as illustrated in Fig. \ref{fig:2.1}. Nonlinear transformations of the group feature space are realized through interactions between group features, thus enhancing nonlinear capabilities.
\begin{figure}[!htbp]
\begin{center}

   \includegraphics[width=0.76\linewidth]{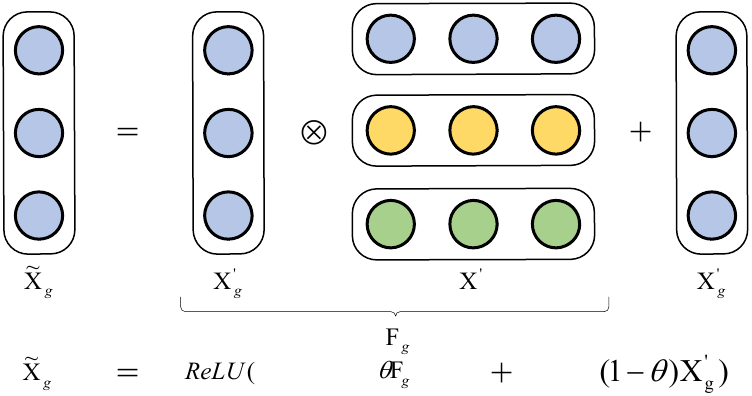}
\end{center}
   \caption{Cross-group feature fusion visualization.}
\label{fig:2.1}

\end{figure}

We apply sum pooling to aggregate the group features after interaction, which can be formulated as
\begin{equation}
{\rm{F}_\emph{g} =\sum\limits_{\emph{j} = 1}^\emph{J=G} {\rm{X_ \emph{g}^{'}}} \otimes  {{\rm{X}}_\emph{j}^{'}} 
 ,\emph{j} = 1,2,...\emph{J} }.
   \end{equation}

If the correlation between two group feature vectors ${{\rm{X}}_g^{'}}$, ${{\rm{X}}_j^{'}}$ is stronger, the product is larger; conversely, if the correlation is weaker, the product is smaller. Upon obtaining ${\rm{F}_\emph{g}}$, we can update the feature ${\rm{ X^{'}}_\emph{g}} \rightarrow {{\rm{\tilde X}}}_\emph{g}$ of each group. The process can be formulated as 
\begin{equation}
   {{\rm{\tilde X}}}_g = ReLU(\theta {\rm{F}_\emph{g}}+(1-\theta){{\rm{X^{'}}_\emph{g}}}),
   \end{equation} 
where $\theta \in [0,1]$ is a constant. Our objective is to predict $p_i$ for each attribute $i$, which can be formulated as
\begin{equation}
    p_i = Softmax({\varphi _i}({\rm{\tilde X}_\emph{g}})),i=1,2,...K,
\end{equation}
where $\emph{Softmax}$ denotes the activation function, $\varphi$ denotes binary classifier.

\subsection{Loss Function}
FocalLoss \citep{lin2017focal} significantly prioritizes the training of challenging samples, thereby improving the efficacy of the balanced cross-entropy loss function. This prioritization is designed to address the imbalance in datasets by focusing more on hard examples that are harder to classify. FocalLoss can be formulated as
\begin{equation}
L_f(p, y) = \sum_{\emph{i}=1}^{K} \left(-\alpha (1 - p_\emph{i})^{\gamma} \log(p_i)\right),
\end{equation}
where $p$ represents the predicted value, $y$ represents the true value, $\alpha$ represents the adjustment category weight that scales the overall positive sample weight, and $\gamma$ represents the weighting factor that adjusts the emphasis on hard and easy samples.

To facilitate the model's ability to learn generalizable features, we employ Kullback-Leibler (KL) divergence to enhance the differentiation among feature spaces learned by task-specific modules. The larger the difference between the distributions of vectors $X$ and $Y$, the larger their product will be. As the difference in distribution increases, the product tends to approach zero. Therefore, we subtracted the KL divergence value to increase the heterogeneity of group features, thereby improving generalization performance. This approach ensures a broader and more distinct representation of features, aiding in the effective generalization across tasks. KL divergence can be formulated as
\begin{equation}
KL({\rm{X||Y}}) = \sum\limits_{n = 0}^{N = size(\rm{X})} {{\rm{X_n}}\log (\frac{{\rm{X_n}}}{Y_n})} 
\end{equation}

The overall loss function can be defined as
\begin{equation}
L = L_f(p, y) - \eta \sum\limits_{{\rm{i}} = 1}^G {\sum\limits_{j = 1,j \ne i}^G KL(\rm{\tilde X}_i,\rm{\tilde X}_j)} .
\end{equation}

\subsection{Dynamic Weighting Strategy}
\label{sec:Dynamic Weighting Strategy}

Building on the concepts of DWA \citep{liu2019end}, we introduce the DWS. DWS uniquely adjusts the weight of each task dynamically based on their respective losses, eliminating the need for multiple iterations to initialize weighted parameters as seen in earlier approaches. Furthermore, DWS accounts for the gradient descent velocity and loss magnitude of each task, offering a more nuanced adaptation to task-specific learning requirements, which can be formulated as
\begin{equation}
\centering
     {\lambda _i}(e) = \frac{K}{{{\rm{1 + }}\beta }}(\frac{{{\varepsilon_i}(e - 1)}}{{\sum\limits_{\emph{i}=1}^K {{\varepsilon_i}(e - 1)} }} + \beta \frac{{{L_i}(e - 1)}}{{\sum\limits_{\emph{i}=1}^K {{L_i}(e - 1)} }}), 
\end{equation}
\begin{equation}
\centering
     {\varepsilon_i}(e - 1) = \frac{{{L_i}(e - 1)}}{{{L_i}(e - 2)}},
\end{equation}
where $K$ represents the number of tasks, $\beta$ represents the relative scale of the loss, ${\varepsilon_i}(e)$ represents the relative gradient descent speed, $\mathit{e}=1,2,...max\_epoch$ represents the training round index (When $\mathit{e}$=1, initialize $\lambda$ with 1.), and ${L_i}(e)$ represents the loss value for task $\emph{i}$ during round $e$ of the training process.

As shown in Fig. \ref{fig:3}, we use the DWA and DWS to conduct experiments on the CelebA dataset and visualize the FocalLoss loss change process before the unweighting of Gray Hair and Oval Face. In the DWA algorithm, the gradient of the easy task decreases faster and the weights will be higher. However, the easy task has smaller loss and higher accuracy, which is lower for the overall model performance improvement. Moreover, setting smaller weights for challenging tasks in the DWA algorithm can lead the model to get stuck in local optima. Therefore, dynamic weighting loss should consider both gradient descent speed and the scale of loss to improve the weighting value of hard tasks.
\begin{figure*}[!htbp]
\begin{center}

   \includegraphics[width=1.0\linewidth]{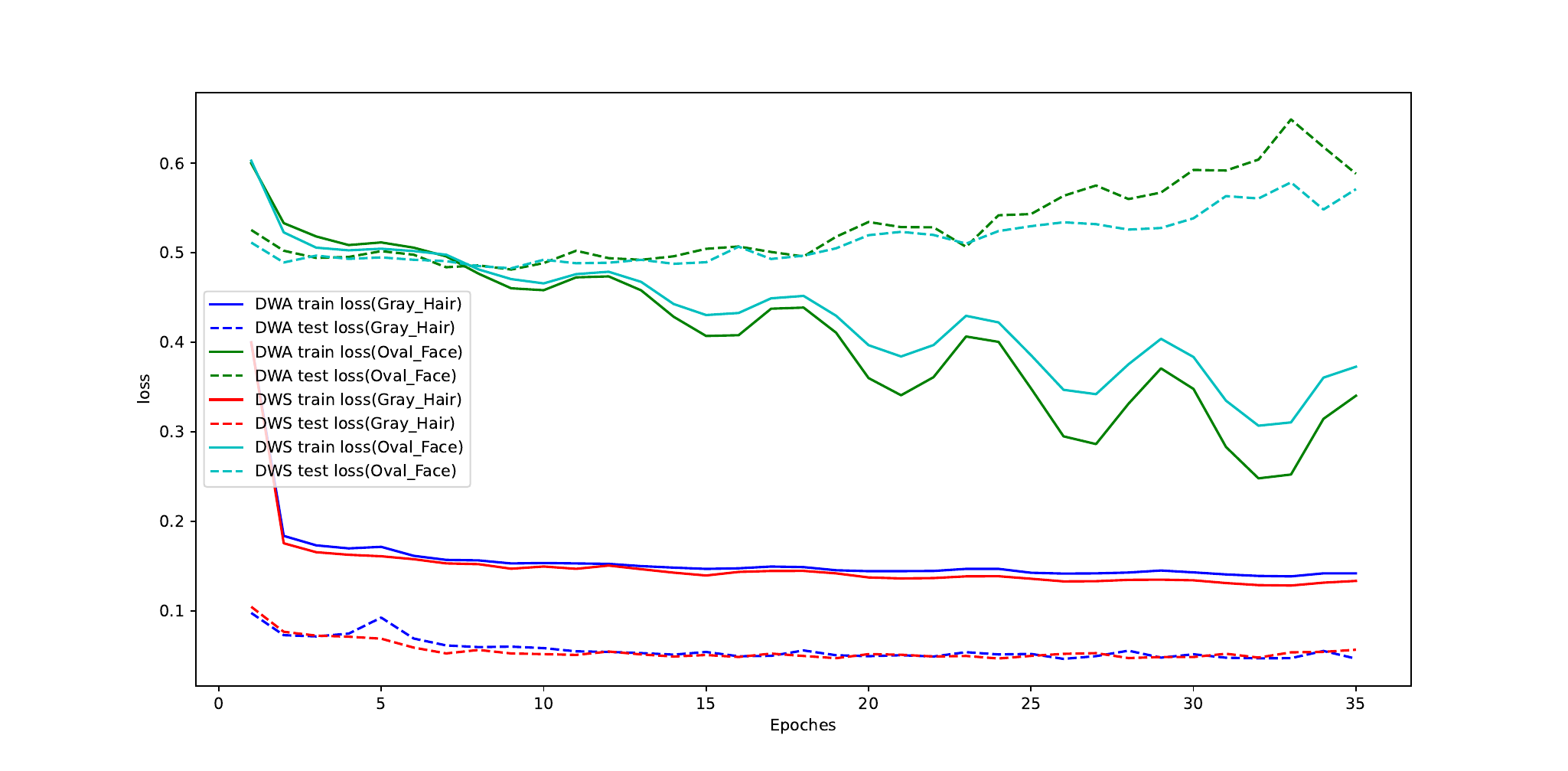}
\end{center}
   \caption{Comparing the train and test losses of Dynamic Weighting Average and Dynamic Weighting Strategy.}
\label{fig:3}

\end{figure*}

\section{Experiment}
\subsection{Datasets and Attribute Grouping}

{\bf Datasets.} 
1) {\bf CelebA}: Developed by the Chinese University of Hong Kong, the CelebA dataset is a comprehensive collection of face images characterized by diverse backgrounds and significant variations in pose. Each image in this dataset is annotated with 40 distinct facial attributes.
2) {\bf LFWA}: Originating from the LFW (Labeled Faces in the Wild) collection, the LFWA dataset compiles 13,233 face images, each with dimensions of 250×250 pixels. This dataset encompasses a wide range of poses, including extreme facial expressions and occlusions. The LFWA, a derivative of the LFW dataset, assigns 40 binary facial attribute labels to each image, consistent with the attributes defined in the CelebA dataset.

{\bf Attribute Grouping.} 
The utilization of grouping methods to streamline MTN models has been a common practice among researchers.
Innovations such as APS \citep{chen2023learning} and MGG-Net \citep{chen2021improving} have refined these methods by focusing on both partial and global facial regions of interest for attribute features. 
Building upon the insights from MGG-Net \citep{chen2021improving} and APS \citep{chen2023learning}, our approach advances the grouping method by investigating facial attributes through both partial and global regions of interest. As shown in Table \ref{tab:1}, we categorize the partial regions of concern into 6 groups based on their locations, including Mouth, Lower Face, Cheeks, Nose, Eyes, Hair), in addition to a seventh group for overarching global regions (Global), culminating in a total of seven distinct groups.
\begin{table}[!htbp]
 \caption{Group of 40 face attributes.}
   \centering
   
     \begin{tabular}{p{4.45em}<{\centering}p{26.55em}<{\centering}}
      \hline
      Group &{Attributes} \\
     \hline
     Mouth & 5 o' Clock Shadow, Big Lips, Mouth Slightly Open, Mustache, Wearing Lipstick, No Beard \\
     Lower Face & Double Chin, Goatee, Wearing Necklace, Wearing Necktie \\
     Cheeks & High Cheekbones, Rosy Cheeks, Sideburns, Wearing Earrings \\
     Nose  & Big Nose, Pointy Nose \\
     Eyes  & Arched Eyebrows, Bags Under Eyes, Bushy Eyebrows, Narrow Eyes, Eyeglass \\
     Hair  & Bald, Bangs, Black Hair, Blond Hair, Brown Hair, Gray Hair, Receding Hairline, Straight Hair, Wavy Hair, Wearing Hat \\
     Global & Attractive, Blurry, Chubby, Heavy Makeup, Male, Oval Face, Pale Skin, Smiling, Young \\
     \hline
     \end{tabular}%
    
   \label{tab:1}%
 \end{table}%
 
\subsection{Parameter Settings}
To facilitate the training of the FAR-AMTN model, the original training and testing partitions of the CelebA and LFWA datasets were employed. Data augmentation techniques, including horizontal flipping and random rotation, were applied to enhance robustness of the model .
The size of the input image is set to 224$\times$224$\times$3. For the Stochastic Gradient Descent (SGD) algorithm, the learning rate is set to 0.001. A batch size of 48 is utilized, and dropout is applied to the fully connected layer with a ratio of 0.15. To maintain consistency across evaluations, all MTN models adhered to a uniform architecture and configuration, as illustrated in Fig. \ref{fig:24}. The hyper-parameters for the CGFF module and the loss function—specifically--$\theta$, $\beta$, and $\eta$--were set to 0.3, 0.5, 0.0025, respectively. Each experiment conducted trained the model from scratch, employing an end-to-end approach.

\subsection{Comparison with State-of-the-art Methods} 
Experiments were conducted to assess the performance of the FAR-AMTN model on the CelebA and LFWA datasets, comparing it against various multi-label \citep{liu2015deep,sharma2020slim,zheng2022general} with multi-task methods \citep{rudd2016moon,lu2017fully,hand2018doing,cao2018partially,mao2020deep,yang2020hierarchical,chen2021improving,shu2021learning,chen2023learning}. The test results are shown in Table \ref{tab:4}.
The results demonstrate that FAR-AMTN outperforms existing methods, achieving accuracy rates of 92.40\% and 87.72\% on the CelebA and LFWA datasets, respectively.
With ResNet50 as the backbone, the parameter count is remarkably low, at only 27.36M.
And it achieves real-time prediction, processing 29.7\% more frames per second than MOON, while reducing memory usage by 80.2\% compared to MOON.
\begin{sidewaystable}[!htbp]
 \caption{Comparison of experimental results between FAR-AMTN and the latest method. On the CelebA and LFWA datasets, respectively.}   
   \centering
   
     \begin{tabular}{ccccccc}
     \hline
            \multirow{2}{*}{Method} & \multicolumn{2}{c}{Accuracy/\%}& \multirow{2}{*}{Backbone}& \multirow{2}{*}{Parameters/×$10^6$} &\multirow{2}{*}{Memory/MB}&\multirow{2}{*}{Test Speed/fps} \\\cline{2-3}
        \multicolumn{1}{r}{} &\multicolumn{1}{p{2.75em}}{CelebA} & \multicolumn{1}{p{2.75em}}{LFWA} & \multicolumn{1}{p{0.10em}}{}{} \\
\hline
 LNets+ANet \citep{liu2015deep}  &87.33&83.85&-& $\textgreater$100&-  \\
  Slim-CNN \citep{sharma2020slim}  &91.24&84.00&-& 0.6&7.9 &\\
  FaRL \citep{zheng2022general} &91.88&86.69&ViT-B&-&-\\
  \hline
  MOON \citep{rudd2016moon} &90.94&-&-&138&457& $\approx$ 33\\
      Branch-64-1.0 \citep{lu2017fully}  &91.26&83.24 &VGG16&4.99&-&15.2 \\
      PS-MCNN-LC \citep{cao2018partially} &92.22 & 87.36 &-&16&-&- \\
      AttCNN \citep{hand2018doing} &90.97 &73.03&-&$\textless$ 6&-&- \\
            DMM-CNN \citep{mao2020deep}  &91.70 &86.56 &ResNet50&360&-&- \\    
                  HFE \citep{yang2020hierarchical} &92.17 & -&ResNet50&-&-&- \\
           SSPL \citep{shu2021learning} &91.77 &86.53 &ResNet50&-&-&- \\
           MGG-Net \citep{chen2021improving}  &92.00 & 87.20 &ResNet50&12.8&-&-\\
            
           ICKD \citep{10064142} &92.02 &86.77&ResNet18&12.32/27.79&-&- \\
           APS \citep{chen2023learning} &92.12&86.74 &VGG16&-&-&-\\  
            FAR-AMTN &\textbf{92.40} &\textbf{87.72} &ResNet50&27.36&90.7&42.8 \\
           \hline
     \end{tabular}

   \label{tab:4}
 \end{sidewaystable}

LNets+ANet, Slim-CNN, and FaRL utilize a single network architecture for attribute recognition, adjusting the predicted outcomes for each attribute with predefined thresholds, achieving accuracies of 87.33\%, 91.24\%, and 91.88\% on the CelebA dataset and 83.55\%, 84.00\%, and 86.69\% on the LFWA dataset, respectively. However, setting these thresholds too high or too low can lead to network instability.

MOON employs 40 independent classifiers to predict each attribute, mapping shared features to various sample spaces, achieving an average accuracy of 90.94\% on the CelebA dataset. Yet, its performance is compromised by the absence of heterogeneous feature learning.

Branch-64-1.0, DMM-CNN, and APS have enhanced model performance through the utilization of shared low-level features and task-specific sub-networks for simultaneous multiple attribute predictions, achieving accuracies of 91.26\%, 91.70\%, and 92.12\% on the CelebA dataset and 83.24\%, 86.56\%, and 86.74\% on the LFWA dataset, respectively. These methods, however, exhibit limited interaction among high-level features.

In contrast, FAR-AMTN advances the MTN framework by incorporating high-level modules with non-shared parameters for feature interaction, thereby improving generalization capabilities while maintaining a lower parameter count. Similar to AttCNN, HFE, and PS-MCNN-LC, FAR-AMTN also employs weighted loss to boost model performance. Moreover, it considers both the speed of gradient descent and the scale of loss, ensuring synchronous task convergence and further performance enhancements. FAR-AMTN surpasses AttCNN, HFE, and PS-MCNN-LC by 1.57\%, 0.25\%, and 0.20\% in accuracy on the CelebA dataset respectively, and FAR-AMTN surpasses AttCNN and PS-MCNN-LC achieves respective improvements of 20.12\%, and 0.41\% on the LFWA dataset.
Unlike MGG-Net, which independently generates predictions using multiple scales of feature maps and then averages them, a process that can degrade average performance and significantly increase the parameter size, FAR-AMTN adopts a more integrated and efficient approach. FAR-AMTN surpasses MGG-Net by 0.43\% and 0.60\% in accuracy on the CelebA and LFWA datasets, respectively.

  \subsection{Ablation Studies}
To determine the efficacy of our proposed method, ablation studies were conducted on the CelebA dataset to assess the impact of the WSGSA, CGFF, and DWS components. These studies utilized a common underlying network architecture with a distinct non-shared high-level network as the backbone. The experiments were structured into four categories: First, an evaluation of the WSGSA module with and without shared parameters was performed. Second, a visual analysis was conducted on the mean and variance of group features processed by CGFF. Third, the implementation and efficacy of DWS were tested. Finally, the FAR-AMTN model was applied to visually identify and analyze facial regions pertinent to each attribute.

The experimental results, presented in Table \ref{tab:l3}, validate the efficacy of the different modules in enhancing the recognition performance of FAR-AMTN. The ResNet50-baseline serves as a reference model, utilizing ResNet50 as the backbone network for shared feature extraction, without grouping strategy. The baseline model uses 40 one-layer fully connected layer classifiers to get attribute prediction results on global features.
\begin{table}[!htbp]
  \caption{Ablation experiments on the CelebA dataset. WSGSA represents Weight-Shared Group-Specific Attention, CGFF represents Cross-Group Feature Fusion, and DWS represents Dynamic Weighting Strategy.}
   \centering
     \begin{tabular}{ccccc}
      \hline
     Method & WSGSA & CGFF & DWS & Accuracy/\% \\
     \hline
     ResNet50-baseline&  &  &  & 91.03 \\
     ResNet50-WSGSA & \checkmark &    &  & 91.50 \\
     ResNet50-CGFF &  & \checkmark &  & 91.86 \\
     ResNet50-DWS &  &  & \checkmark & 91.85 \\
     \hline
     ResNet50-WSGSA-CGFF & \checkmark & \checkmark &  & 92.17 \\
     ResNet50-WSGSA-DWS & \checkmark &  & \checkmark & 91.90 \\
     FAR-AMTN & \checkmark & \checkmark & \checkmark & \textbf{92.40} \\
     \hline
     \end{tabular}%
   
   \label{tab:l3}%
 \end{table}%

Upon the integration of the WSGSA module into ResNet50 (ResNet50-WSGSA), there was a notable improvement in accuracy by 0.47\%. Furthermore, the incorporation of both the Cross-Group Feature Fusion (ResNet50-CGFF) and Dynamic Weighting Strategy (ResNet50-DWS) modules significantly enhances the performance of the FAR-AMTN model. Remarkably, the synergistic combination of these three modules leads to state-of-the-art recognition performance, with FAR-AMTN achieving an improvement in accuracy of 1.50\% over the ResNet50 baseline.

  {\bf Effectiveness of Weight-Shared Group-Specific Attention.}
In order to evaluate the effectiveness of the WSGSA module with both shared and non-shared weights regarding recognition performance and parameter efficiency, experiments were conducted focusing on channel attention, spatial attention \citep{jaderberg2015spatial}, and non-local attention \citep{wang2018non}.
The weight-sharing mechanism utilized in the WSGSA module, as depicted in Fig. \ref{fig:4}(b), employs three convolutional layers with shared parameters for learning group features, complemented by an attention layer without shared parameters to further refine the representation of group features.

\begin{figure}[!htbp]
\begin{center}

   \includegraphics[width=1.0\linewidth]{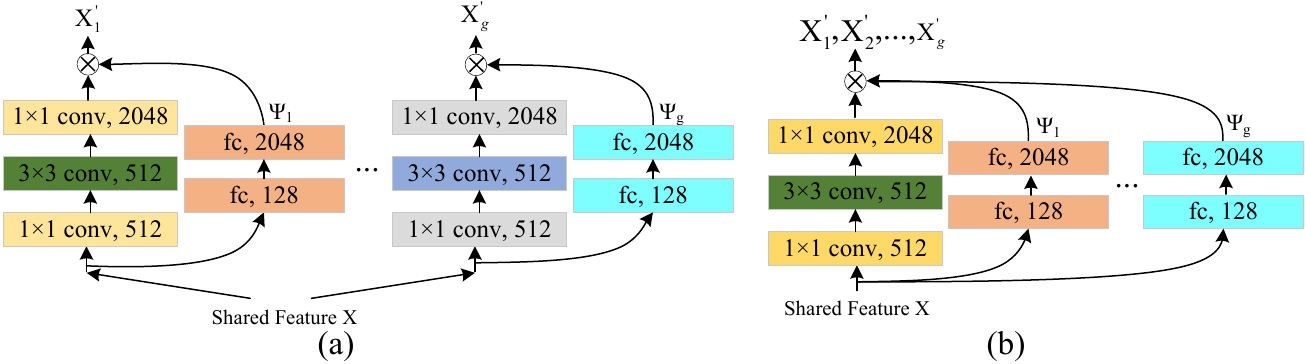}
\end{center}
   \caption{Weight-sharing mechanism for WSGSA module. (a) The method with non-shared weights, where both the convolutional layer and the attention layer parameters are not shared; (b) Weight-sharing mechanism, where the convolutional layer parameters are shared while the attention layer parameters are not shared.}
\label{fig:4}

\end{figure}

As shown in Table \ref{tab:11}, relative to the baseline, the increases in parameter counts for channel attention, spatial attention, and non-local attention utilizing shared weights were 3.67M, $\approx$ 0M, and 58.72M, respectively.

In contrast, when compared to approaches without a parameter sharing mechanism, the parameter counts for channel attention, spatial attention, and non-local attention in the shared parameter framework were reduced by 23.88\%, 27.40\%, and 9.79\%, respectively. It is important to note that channel attention and non-local attention experienced enhancements in performance, whereas spatial attention saw a marginal reduction. This reduction in spatial attention is attributed to the limited scale of spatial features within the WSGSA module, which harbors less information, leading to a slight performance decrement. These findings imply that the parameter-sharing mechanism effectively augments the representational capabilities of group features without a substantial increase in parameter count.
\begin{table}[!htbp]
  \caption{The WSGSA is compared with the different parameter-sharing methods in the CelebA dataset. WNGSA denotes weight non-shared group-specific attention mechanism; WSGSA denotes weight shared group-specific attention mechanism; CA denotes channel attention; SA denotes spatial attention; NL denotes non-local attention.}
   \centering
  
     \begin{tabular}{ccc}
      \hline
Method &Parameters/×$10^{6}$& Accuracy/\% \\
     \hline
     baseline& 23.69 & 91.03 \\
     WNGSA-CA& 36.30 & 91.41 \\
     WNGSA-SA& 32.63 & 91.26 \\
     WNGSA-NL& 91.35 & 91.39 \\
     \hline
     WSGSA-CA& 27.36  & \textbf{91.50} \\
     WSGSA-SA& 23.69  & 91.22 \\
     WSGSA-NL& 82.41  & 91.45 \\
     \hline
     \end{tabular}%
    
   \label{tab:11}%
 \end{table}%

  {\bf Effectiveness of Cross-Group Feature Fusion.} To demonstrate the effectiveness of CGFF, we computed the group features ${{\rm{\tilde X}}}_{g},g=1,2,...\emph{G}$ and depicted the distributions of their means and variances as shown in \ref{fig:5}. It can be observed that the variance distribution of group features using the CGFF module (see Fig. \ref{fig:5}(b)) is more concentrated compared to the distribution of group features without the CGFF module (see Fig. \ref{fig:5}(a)). Additionally, there is a noticeable increase in variability among different groups concerning both mean and variance distributions. This observation implies that with the application of CGFF, individual group features are more aligned towards the central value within the CGFF module. Nevertheless, it is essential to acknowledge the significant diversity in the distributions derived from different groups, which aligns with the fundamental principles of the grouping method.
\begin{figure}[!htbp]
\begin{center}

   \includegraphics[width=0.96\linewidth]{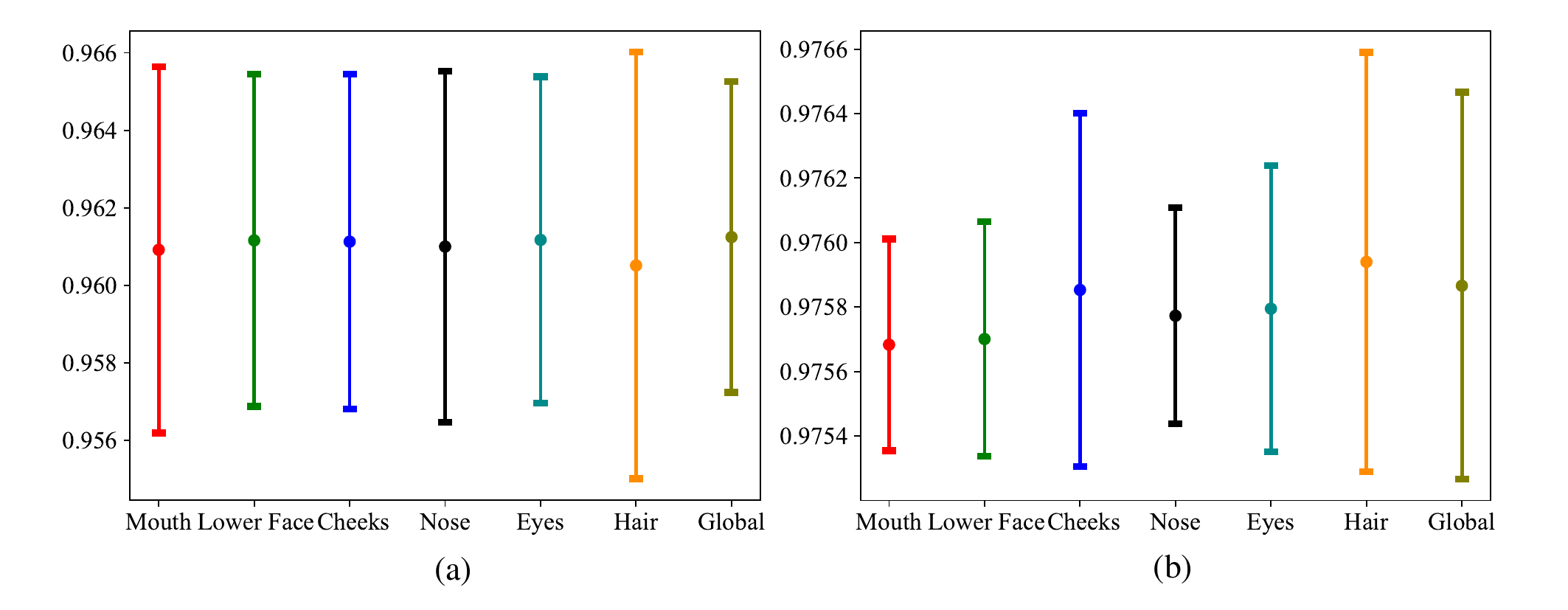}
\end{center}
   \caption{Analysis of the mean and variance of group features obtained with and without CGFF. (a) Analysis of the mean and variance of the features learned without CGFF; (b) Analysis of the mean and variance of the features learned with CGFF.}
\label{fig:5}

\end{figure}
To assess the effect of varying $\theta$ values within the range [0,1] on the performance of the CGFF module, we executed experiments with different $\theta$ values for the CGFF module. The impact of these different values can be observed in Table \ref{tab:11l}. When $\theta$=0, no feature interaction occurs, resulting in an accuracy of only 91.41\%. As $\theta$ gradually increases, the interacted features begin to dominate. When $\theta$ is set to 0.6, the accuracy increases by 0.91\% compared to $\theta$ = 0.  This indicates that employing CGFF enhances group feature embedding through feature interaction, subsequently improving model performance.
 \begin{table}[htbp]
  \caption{Comparison of recognition accuracy obtained by different $\theta$ values in CGFF.}
   \centering
     \begin{tabular}{ccccccc}
      \hline
$\theta$ &0 &0.1 & 0.4 & 0.6 &0.8 &1.0\\
    Accuracy/\%& 91.03& 91.73 & 91.79& \textbf{91.86} & 91.61& 91.49 \\
     \hline
     \end{tabular}%
    
   \label{tab:11l}%
 \end{table}%

 {\bf Effectiveness of Dynamic Weighting Strategy.}
To ascertain the efficacy of the DWS within the MTN framework, we evaluated its influence on performance of proposed method during the training process. Initially, the MTN model, based on a ResNet50 baseline, was trained and subsequently assessed on a testing set. Subsequently, we implemented the DWS technique, trained the ResNet50-DWS model, and performed another evaluation on the testing set.

As shown in Fig. \ref{fig:8}, the implementation of DWS significantly reduced the overall loss, thereby enhancing recognition performance.
Specifically, employing DWS with $\beta=0.5$ resulted in a recognition accuracy of 91.85\% on the CelebA dataset, an improvement of 0.90\% over the baseline method without dynamic weighting, and a 0.1\% increase compared to using DWS with $\beta=0.1$.
\begin{figure}[!htbp]
\begin{center}

  \includegraphics[width=0.96\linewidth]{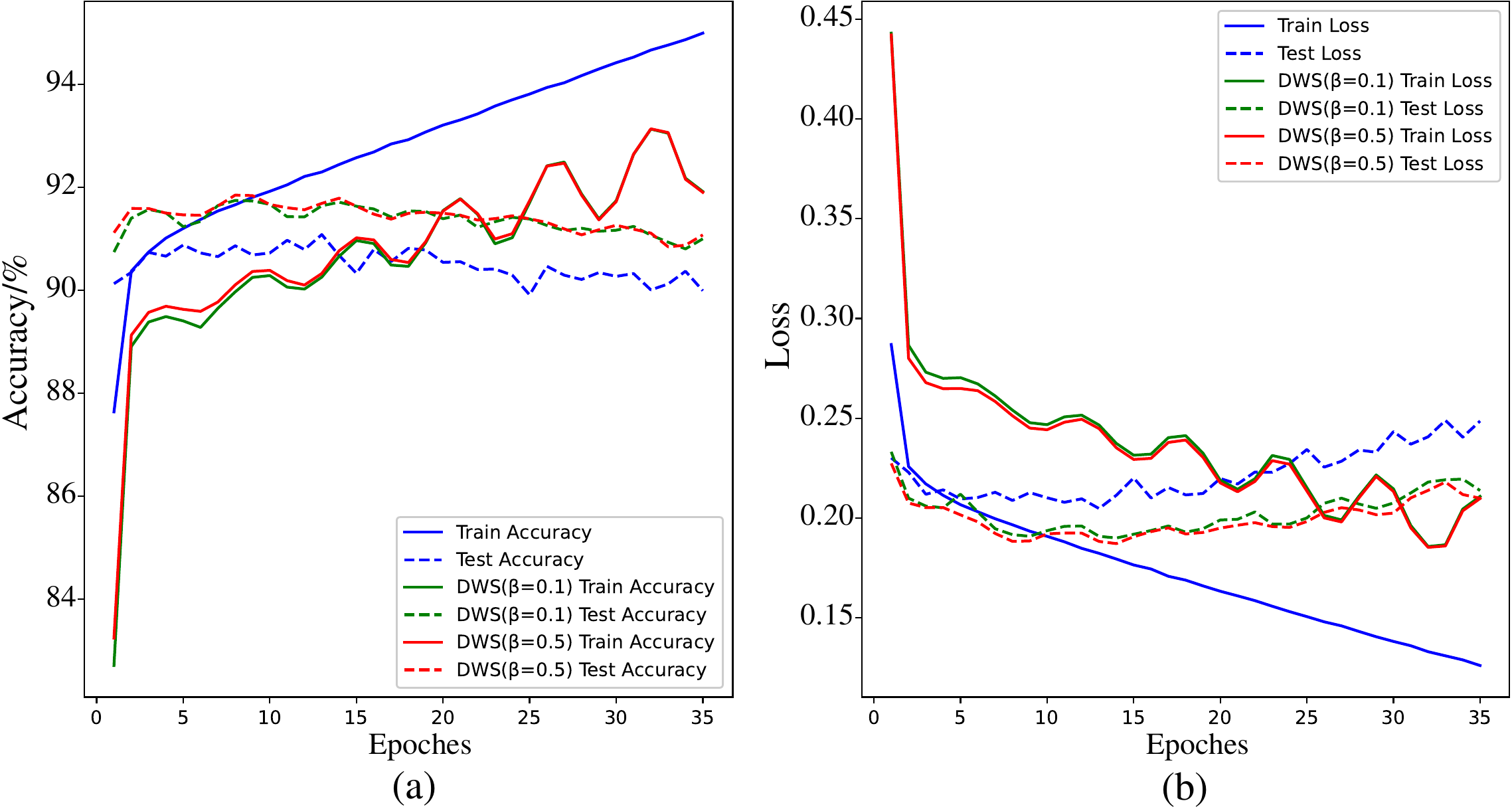}
\end{center}
  \caption{Comparison of accuracy and loss of training and test results obtained using DWS with different $\beta$ values and without DWS on CelebA dataset.}
\label{fig:8}

\end{figure}

Empirical observations indicated a reduction in training loss, albeit with some fluctuations. These fluctuations are believed to be a consequence of integrating DWS, which adjusts both the rate of gradient descent and the scale of loss throughout an epoch, leading to variability in loss figures.

Based on these findings, several conclusions can be drawn: (1) The absence of the WSGSA module significantly impairs the feature extraction efficiency of the MTN, while concurrently increasing the parameter count; (2) Lacking the CGFF module compromises the ability to fully comprehend semantic relationships between facial attributes, potentially detracting from the overall efficacy; (3) The Dynamic Weight Strategy (DWS) is essential for the MTN framework to achieve effective convergence. Without this component, the model may not reach optimal performance levels.

 {\bf Visualization.} To visualize the heatmaps generated by FAR-AMTN, we employed Grad-CAM++ \citep{chattopadhay2018grad}, as shown in Fig. \ref{fig:6}. These heatmaps precisely delineate pertinent facial areas such as the mouth and nose groups, showcasing the precision in focusing on specific facial features. Moreover, the heatmaps effectively minimize focus on facial regions irrelevant to the identified attribute. These visualizations substantiate the capability of proposed method to learn heterogeneous group features and solidify the assertion that it develops robust group feature representations, thereby augmenting overall generalization capacity of the MTN.
  \begin{figure}[!htbp]
   \begin{center}
      \includegraphics[width=0.96\linewidth]{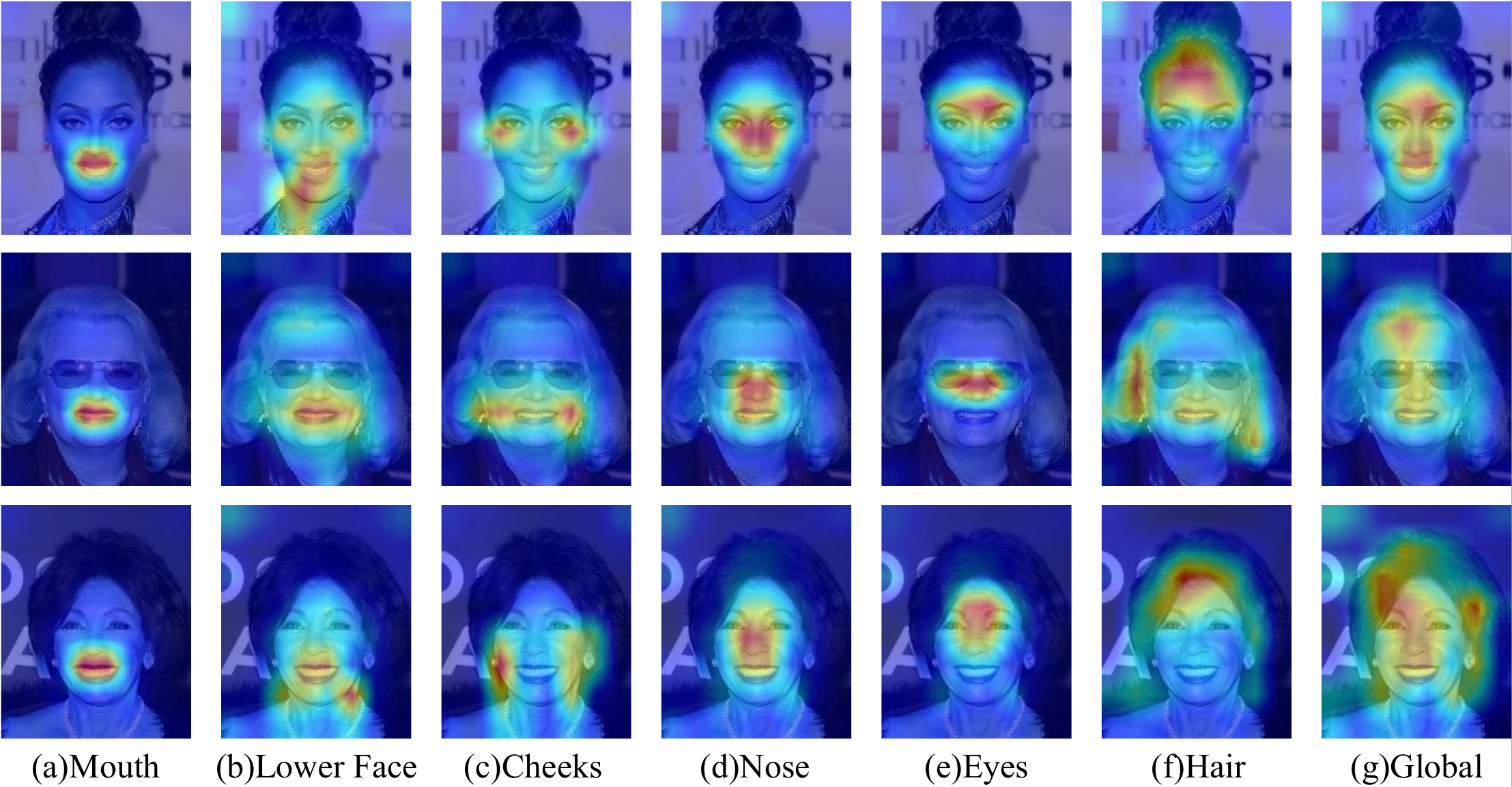}
   \end{center}
      \caption{Heatmaps visualization of FAR-AMTN for attribute recognition.}
   \label{fig:6}
\end{figure}

\section{Conclusion and Future Work}
In this study, we introduce the FAR-AMTN model, a novel approach to FAR. This method focuses on efficiently learning heterogeneous group features while simultaneously reducing the parameter count. The WSGSA module utilizes group-specific modules and a parameter-sharing group-specific attention mechanism to enable efficient feature learning. The CGFF module is instrumental in facilitating feature interactions, thus allowing the FAR-AMTN model to uncover and leverage the semantic relationships among group features. To ensure the model converges effectively, we employ Dynamic Weight Sharing (DWS). Performance evaluations conducted on the CelebA and LFWA datasets illustrate that the FAR-AMTN model outperforms existing methods, showcasing its superiority in FAR tasks while maintaining an efficient and streamlined architecture.

In this study, we concentrate on modeling heterogeneous attention regions across various groups using attention mechanisms, which effectively reduces the model parameters. Our observations reveal that the learning complexities of different tasks vary; concurrently training both simple and complex tasks may result in overfitting for certain tasks. For future research, we aim to utilize multi-task models to delve deeper into the methods of modeling task learning complexity.

\section{Declarations}
\subsection{Funding}
This study was partially supported by National Key Research and Development Program of China (Grant Number 2022YFB330570) and Science and Technology Innovation Plan Of Shanghai Science and Technology Commission (Grant Number 21511104302).
\subsection{Conflicts of interest/Competing interests}
The authors declare that they have no competing interests.
\subsection{Availability of data and material (data transparency)}
The data that has been used is confidential.
\subsection{Code availability (software application or custom code)}
The code used in this study is available upon request to the corresponding author.
\subsection{Authors' contributions}
Gong Gao: Project administration, Conceptualization, Formal analysis, Methodology, Resources, Validation, Visualization, Writing – original draft, Writing – review \& editing. Zekai Wang: Writing – review \& editing. Weidong Zhao: Conceptualization. Xianhui Liu: Conceptualization, Funding acquisition.

\bibliographystyle{elsarticle-harv}
\bibliography{egbib}

@String(AAAI  = {AAAI})

@String(ICME  = {Int. Conf. Multimedia and Expo})

@String(ICME  =	{ICME})

@inproceedings{kou2023character,
  title={Character as pixels: a controllable prompt adversarial attacking framework for black-box text guided image generation models},
  author={Kou, Ziyi and Pei, Shichao and Tian, Yijun and Zhang, Xiangliang},
  booktitle={Proceedings of the Thirty-Second International Joint Conference on Artificial Intelligence (IJCAI-23)},
  year={2023},
  organization={Proceedings of the Thirty-Second International Joint Conference on~…}
}

@inproceedings{zaeemzadeh2021face,
  title={Face image retrieval with attribute manipulation},
  author={Zaeemzadeh, Alireza and Ghadar, Shabnam and Faieta, Baldo and Lin, Zhe and Rahnavard, Nazanin and Shah, Mubarak and Kalarot, Ratheesh},
  booktitle={Proceedings of the IEEE/CVF International Conference on Computer Vision},
  pages={12116--12125},
  year={2021}
}

@inproceedings{li2015multi,
  title={Multi-task model and feature joint learning},
  author={Li, Ya and Tian, Xinmei and Liu, Tongliang and Tao, Dacheng},
  booktitle={Twenty-fourth international joint conference on artificial intelligence},
  year={2015}
}

@inproceedings{lu2017fully,
  title={Fully-adaptive feature sharing in multi-task networks with applications in person attribute classification},
  author={Lu, Yongxi and Kumar, Abhishek and Zhai, Shuangfei and Cheng, Yu and Javidi, Tara and Feris, Rogerio},
  booktitle={Proceedings of the IEEE conference on computer vision and pattern recognition},
  pages={5334--5343},
  year={2017}
}

@inproceedings{chen2021improving,
  title={Improving facial attribute recognition by group and graph learning},
  author={Chen, Zhenghao and Gu, Shuhang and Zhu, Feng and Xu, Jing and Zhao, Rui},
  booktitle={2021 IEEE International Conference on Multimedia and Expo (ICME)},
  pages={1--6},
  year={2021},
  organization={IEEE}
}

@article{mao2020deep,
  title={Deep multi-task multi-label CNN for effective facial attribute classification},
  author={Mao, Longbiao and Yan, Yan and Xue, Jing-Hao and Wang, Hanzi},
  journal={IEEE Transactions on Affective Computing},
  volume={13},
  number={2},
  pages={818--828},
  year={2020},
  publisher={IEEE}
}

@inproceedings{liu2015deep,
  title={Deep learning face attributes in the wild},
  author={Liu, Ziwei and Luo, Ping and Wang, Xiaogang and Tang, Xiaoou},
  booktitle={Proceedings of the IEEE international conference on computer vision},
  pages={3730--3738},
  year={2015}
}

@article{liu2018large,
  title={Large-scale celebfaces attributes (celeba) dataset},
  author={Liu, Ziwei and Luo, Ping and Wang, Xiaogang and Tang, Xiaoou},
  journal={Retrieved August},
  volume={15},
  number={2018},
  pages={11},
  year={2018}
}

@inproceedings{zheng2022general,
  title={General facial representation learning in a visual-linguistic manner},
  author={Zheng, Yinglin and Yang, Hao and Zhang, Ting and Bao, Jianmin and Chen, Dongdong and Huang, Yangyu and Yuan, Lu and Chen, Dong and Zeng, Ming and Wen, Fang},
  booktitle={Proceedings of the IEEE/CVF Conference on Computer Vision and Pattern Recognition},
  pages={18697--18709},
  year={2022}
}

@ARTICLE{10064142,
  author={Chen, Si and Zhu, Xueyan and Yan, Yan and Zhu, Shunzhi and Li, Shao-Zi and Wang, Da-Han},
  journal={IEEE Transactions on Circuits and Systems for Video Technology},
  title={Identity-Aware Contrastive Knowledge Distillation for Facial Attribute Recognition},
  year={2023},
  volume={},
  number={},
  pages={1-1},
  doi={10.1109/TCSVT.2023.3253799}}

@inproceedings{shu2021learning,
  title={Learning spatial-semantic relationship for facial attribute recognition with limited labeled data},
  author={Shu, Ying and Yan, Yan and Chen, Si and Xue, Jing-Hao and Shen, Chunhua and Wang, Hanzi},
  booktitle={Proceedings of the IEEE/CVF Conference on Computer Vision and Pattern Recognition},
  pages={11916--11925},
  year={2021}
}

@article{cao2019learning,
  title={Learning imbalanced datasets with label-distribution-aware margin loss},
  author={Cao, Kaidi and Wei, Colin and Gaidon, Adrien and Arechiga, Nikos and Ma, Tengyu},
  journal={Advances in neural information processing systems},
  volume={32},
  year={2019}
}

@inproceedings{yang2020hierarchical,
  title={Hierarchical feature embedding for attribute recognition},
  author={Yang, Jie and Fan, Jiarou and Wang, Yiru and Wang, Yige and Gan, Weihao and Liu, Lin and Wu, Wei},
  booktitle={Proceedings of the IEEE/CVF conference on computer vision and pattern recognition},
  pages={13055--13064},
  year={2020}
}

@inproceedings{hand2018doing,
title={Doing the best we can with what we have: Multi-label balancing with selective learning for attribute prediction},
author={Hand, Emily and Castillo, Carlos and Chellappa, Rama},
booktitle={Proceedings of the AAAI Conference on Artificial Intelligence},
volume={32},
number={1},
year={2018}
}

@article{chen2023learning,
  title={Learning an attention-aware parallel sharing network for facial attribute recognition},
  author={Chen, Si and Lai, Xinyu and Yan, Yan and Wang, Da-Han and Zhu, Shunzhi},
  journal={Journal of Visual Communication and Image Representation},
  pages={103745},
  year={2023},
  publisher={Elsevier}
}

@inproceedings{wang2019dynamic,
  title={Dynamic curriculum learning for imbalanced data classification},
  author={Wang, Yiru and Gan, Weihao and Yang, Jie and Wu, Wei and Yan, Junjie},
  booktitle={Proceedings of the IEEE/CVF international conference on computer vision},
  pages={5017--5026},
  year={2019}
}

@article{zhang2021deep,
  title={Deep long-tailed learning: A survey},
  author={Zhang, Yifan and Kang, Bingyi and Hooi, Bryan and Yan, Shuicheng and Feng, Jiashi},
  journal={arXiv preprint arXiv:2110.04596},
  year={2021}
}

@inproceedings{park2021influence,
  title={Influence-balanced loss for imbalanced visual classification},
  author={Park, Seulki and Lim, Jongin and Jeon, Younghan and Choi, Jin Young},
  booktitle={Proceedings of the IEEE/CVF International Conference on Computer Vision},
  pages={735--744},
  year={2021}
}

@article{santos2018cross,
  title={Cross-validation for imbalanced datasets: avoiding overoptimistic and overfitting approaches},
  author={Santos, Miriam Seoane and Soares, Jastin Pompeu and Abreu, Pedro Henrigues and Araujo, Helder and Santos, Joao},
  journal={IEEE Computational Intelligence Magazine},
  volume={13},
  number={4},
  pages={59--76},
  year={2018},
  publisher={IEEE}
}

@inproceedings{huang2008labeled,
  title={Labeled faces in the wild: A database forstudying face recognition in unconstrained environments},
  author={Huang, Gary B and Mattar, Marwan and Berg, Tamara and Learned-Miller, Eric},
  booktitle={Workshop on faces in'Real-Life'Images: detection, alignment, and recognition},
  year={2008}
}

@inproceedings{he2016deep,
  title={Deep residual learning for image recognition},
  author={He, Kaiming and Zhang, Xiangyu and Ren, Shaoqing and Sun, Jian},
  booktitle={Proceedings of the IEEE conference on computer vision and pattern recognition},
  pages={770--778},
  year={2016}
}

@inproceedings{he2017adaptively,
  title={Adaptively weighted multi-task deep network for person attribute classification},
  author={He, Keke and Wang, Zhanxiong and Fu, Yanwei and Feng, Rui and Jiang, Yu-Gang and Xue, Xiangyang},
  booktitle={Proceedings of the 25th ACM international conference on Multimedia},
  pages={1636--1644},
  year={2017}
}

@inproceedings{liu2019end,
  title={End-to-end multi-task learning with attention},
  author={Liu, Shikun and Johns, Edward and Davison, Andrew J},
  booktitle={Proceedings of the IEEE/CVF conference on computer vision and pattern recognition},
  pages={1871--1880},
  year={2019}
}

@inproceedings{cao2018partially,
  title={Partially shared multi-task convolutional neural network with local constraint for face attribute learning},
  author={Cao, Jiajiong and Li, Yingming and Zhang, Zhongfei},
  booktitle={Proceedings of the IEEE Conference on computer vision and pattern recognition},
  pages={4290--4299},
  year={2018}
}

@inproceedings{ehrlich2016facial,
  title={Facial attributes classification using multi-task representation learning},
  author={Ehrlich, Max and Shields, Timothy J and Almaev, Timur and Amer, Mohamed R},
  booktitle={Proceedings of the IEEE Conference on Computer Vision and Pattern Recognition Workshops},
  pages={47--55},
  year={2016}
}

@inproceedings{rudd2016moon,
  title={Moon: A mixed objective optimization network for the recognition of facial attributes},
  author={Rudd, Ethan M and G{\"u}nther, Manuel and Boult, Terrance E},
  booktitle={Computer Vision--ECCV 2016: 14th European Conference, Amsterdam, The Netherlands, October 11-14, 2016, Proceedings, Part V 14},
  pages={19--35},
  year={2016},
  organization={Springer}
}

@inproceedings{ding2018deep,
  title={A deep cascade network for unaligned face attribute classification},
  author={Ding, Hui and Zhou, Hao and Zhou, Shaohua and Chellappa, Rama},
  booktitle={Proceedings of the AAAI Conference on Artificial Intelligence},
  volume={32},
  number={1},
  year={2018}
}

@inproceedings{kalayeh2017improving,
  title={Improving facial attribute prediction using semantic segmentation},
  author={Kalayeh, Mahdi M and Gong, Boqing and Shah, Mubarak},
  booktitle={Proceedings of the IEEE Conference on Computer Vision and Pattern Recognition},
  pages={6942--6950},
  year={2017}
}

@article{du2021parameter,
  title={Parameter-free loss for class-imbalanced deep learning in image classification},
  author={Du, Jie and Zhou, Yanhong and Liu, Peng and Vong, Chi-Man and Wang, Tianfu},
  journal={IEEE Transactions on Neural Networks and Learning Systems},
  year={2021},
  publisher={IEEE}
}

@inproceedings{lingenfelter2021improving,
  title={Improving evaluation of facial attribute prediction models},
  author={Lingenfelter, Bryson and Hand, Emily M},
  booktitle={2021 16th IEEE International Conference on Automatic Face and Gesture Recognition (FG 2021)},
  pages={1--7},
  year={2021},
  organization={IEEE}
}

@inproceedings{song2022prior,
  title={Prior-Guided Multi-scale Fusion Transformer for Face Attribute Recognition},
  author={Song, Shaoheng and Huang, Huaibo and Wang, Jiaxiang and Zheng, Aihua and He, Ran},
  booktitle={Pattern Recognition and Computer Vision: 5th Chinese Conference, PRCV 2022, Shenzhen, China, November 4--7, 2022, Proceedings, Part I},
  pages={645--659},
  year={2022},
  organization={Springer}
}

@article{serna2022sensitive,
  title={Sensitive loss: Improving accuracy and fairness of face representations with discrimination-aware deep learning},
  author={Serna, Ignacio and Morales, Aythami and Fierrez, Julian and Obradovich, Nick},
  journal={Artificial Intelligence},
  volume={305},
  pages={103682},
  year={2022},
  publisher={Elsevier}
}

@inproceedings{lin2017focal,
  title={Focal loss for dense object detection},
  author={Lin, Tsung-Yi and Goyal, Priya and Girshick, Ross and He, Kaiming and Doll{\'a}r, Piotr},
  booktitle={Proceedings of the IEEE international conference on computer vision},
  pages={2980--2988},
  year={2017}
}

@article{lv2022synchronous,
  title={A synchronous detection-segmentation method for oversized gangue on a coal preparation plant based on multi-task learning},
  author={Lv, Ziqi and Wang, Weidong and Zhang, Kanghui and Li, Wujin and Feng, Junda and Xu, Zhiqiang},
  journal={Minerals Engineering},
  volume={187},
  pages={107806},
  year={2022},
  publisher={Elsevier}
}

@article{jaderberg2015spatial,
  title={Spatial transformer networks},
  author={Jaderberg, Max and Simonyan, Karen and Zisserman, Andrew and others},
  journal={Advances in neural information processing systems},
  volume={28},
  year={2015}
}

@inproceedings{wang2018non,
  title={Non-local neural networks},
  author={Wang, Xiaolong and Girshick, Ross and Gupta, Abhinav and He, Kaiming},
  booktitle={Proceedings of the IEEE conference on computer vision and pattern recognition},
  pages={7794--7803},
  year={2018}
}

@inproceedings{sharma2020slim,
  title={Slim-cnn: A light-weight cnn for face attribute prediction},
  author={Sharma, Ankit Kumar and Foroosh, Hassan},
  booktitle={2020 15th IEEE International Conference on Automatic Face and Gesture Recognition (FG 2020)},
  pages={329--335},
  year={2020},
  organization={IEEE}
}

@inproceedings{chattopadhay2018grad,
  title={Grad-cam++: Generalized gradient-based visual explanations for deep convolutional networks},
  author={Chattopadhay, Aditya and Sarkar, Anirban and Howlader, Prantik and Balasubramanian, Vineeth N},
  booktitle={2018 IEEE winter conference on applications of computer vision (WACV)},
  pages={839--847},
  year={2018},
  organization={IEEE}
}

@inproceedings{iranmanesh2018deep,
  title={Deep sketch-photo face recognition assisted by facial attributes},
  author={Iranmanesh, Seyed Mehdi and Kazemi, Hadi and Soleymani, Sobhan and Dabouei, Ali and Nasrabadi, Nasser M},
  booktitle={2018 IEEE 9th International Conference on Biometrics Theory, Applications and Systems (BTAS)},
  pages={1--10},
  year={2018},
  organization={IEEE}
}

@article{pattnaik2023face,
  title={A face recognition taxonomy and review framework towards dimensionality, modality and feature quality},
  author={Pattnaik, Ipsita and Dev, Amita and Mohapatra, AK},
  journal={Engineering Applications of Artificial Intelligence},
  volume={126},
  pages={107056},
  year={2023},
  publisher={Elsevier}
}

@article{maroto2023active,
  title={Active learning based on computer vision and human--robot interaction for the user profiling and behavior personalization of an autonomous social robot},
  author={Maroto-G{\'o}mez, Marcos and Marqu{\'e}s-Villaroya, Sara and Castillo, Jos{\'e} Carlos and Castro-Gonz{\'a}lez, {\'A}lvaro and Malfaz, Mar{\'\i}a},
  journal={Engineering Applications of Artificial Intelligence},
  volume={117},
  pages={105631},
  year={2023},
  publisher={Elsevier}
}

@article{yang2023dfsgan,
  title={DFSGAN: Introducing editable and representative attributes for few-shot image generation},
  author={Yang, Mengping and Niu, Saisai and Wang, Zhe and Li, Dongdong and Du, Wenli},
  journal={Engineering Applications of Artificial Intelligence},
  volume={117},
  pages={105519},
  year={2023},
  publisher={Elsevier}
}

@article{liu2023sketch2photo,
  title={Sketch2Photo: Synthesizing photo-realistic images from sketches via global contexts},
  author={Liu, Heng and Xu, Yao and Chen, Feng},
  journal={Engineering Applications of Artificial Intelligence},
  volume={117},
  pages={105608},
  year={2023},
  publisher={Elsevier}
}

@article{tao2024hierarchical,
  title={Hierarchical attention network with progressive feature fusion for facial expression recognition},
  author={Tao, Huanjie and Duan, Qianyue},
  journal={Neural Networks},
  volume={170},
  pages={337--348},
  year={2024},
  publisher={Elsevier}
}

@inproceedings{ye2023taskexpert,
  title={Taskexpert: Dynamically assembling multi-task representations with memorial mixture-of-experts},
  author={Ye, Hanrong and Xu, Dan},
  booktitle={Proceedings of the IEEE/CVF International Conference on Computer Vision},
  pages={21828--21837},
  year={2023}
}

@inproceedings{ye2022taskprompter,
  title={Taskprompter: Spatial-channel multi-task prompting for dense scene understanding},
  author={Ye, Hanrong and Xu, Dan},
  booktitle={The Eleventh International Conference on Learning Representations},
  year={2022}
}

@inproceedings{vandenhende2020mti,
  title={Mti-net: Multi-scale task interaction networks for multi-task learning},
  author={Vandenhende, Simon and Georgoulis, Stamatios and Van Gool, Luc},
  booktitle={Computer Vision--ECCV 2020: 16th European Conference, Glasgow, UK, August 23--28, 2020, Proceedings, Part IV 16},
  pages={527--543},
  year={2020},
  organization={Springer}
}

@inproceedings{ye2022inverted,
  title={Inverted pyramid multi-task transformer for dense scene understanding},
  author={Ye, Hanrong and Xu, Dan},
  booktitle={European Conference on Computer Vision},
  pages={514--530},
  year={2022},
  organization={Springer}
}

@inproceedings{xu2022mtformer,
  title={Mtformer: Multi-task learning via transformer and cross-task reasoning},
  author={Xu, Xiaogang and Zhao, Hengshuang and Vineet, Vibhav and Lim, Ser-Nam and Torralba, Antonio},
  booktitle={European Conference on Computer Vision},
  pages={304--321},
  year={2022},
  organization={Springer}
}

@article{tao2023smoke,
  title={Smoke Recognition in Satellite Imagery via an Attention Pyramid Network With Bidirectional Multi-Level Multi-Granularity Feature Aggregation and Gated Fusion},
  author={Tao, Huanjie},
  journal={IEEE Internet of Things Journal},
  year={2023},
  publisher={IEEE}
}

\end{sloppy}  
\end{document}